\documentclass{article} 
\usepackage[preprint]{colm2026_conference}

\usepackage{microtype}
\usepackage{hyperref}
\usepackage{url}
\usepackage{booktabs}
\usepackage{amsmath}
\usepackage{graphicx}

\usepackage{lineno}

\definecolor{darkblue}{rgb}{0, 0, 0.5}
\hypersetup{colorlinks=true, citecolor=darkblue, linkcolor=darkblue, urlcolor=darkblue}

\title{Reward Weighted Classifier-Free Guidance as Policy Improvement in Autoregressive Models}

\author{Alexander Peysakhovich \thanks{Sutter Hill Ventures} \\
\And
William Berman \thanks{Sutter Hill Ventures}
}

\begin{document}

\ifcolmsubmission
\linenumbers
\fi

\maketitle

\begin{abstract}
Consider an autoregressive model that produces outputs $x$ (e.g., answers to questions, molecules) each of which can be summarized by an attribute vector $y$ (e.g., helpfulness vs. harmlessness, or bio-availability vs. lipophilicity). An arbitrary reward function $r(y)$ encodes tradeoffs between these properties. Typically, tilting the model's sampling distribution to increase this reward is done at training time via reinforcement learning. However, if the reward function changes, re-alignment requires re-training. In this paper, we show that a reward weighted classifier-free guidance (RCFG) can act as a policy improvement operator in this setting, approximating tilting the sampling distribution by the $Q$ function. We apply RCFG to molecular generation, demonstrating that it can optimize novel reward functions at test time. Finally, we show that using RCFG as a teacher and distilling into the base policy to serve as a warm start significantly speeds up convergence for standard RL.
\end{abstract}

\section{Introduction}
A fundamental practical problem in modern machine learning is tilting or `aligning' a base generative model $\pi$ to a reward function $r$. Alignment is typically performed via the use of reinforcement learning to construct a new model $\hat{\pi}$ that outputs responses that score high on $x$ while regularizing the policy to stay close to $\pi$ \citep{ouyang2022training, ziegler2019fine,schulman2017proximal,shao2024deepseekmath,rafailov2023direct}.

In many cases, reward models can change across users, use-cases, or time. In language model alignment, the desired tradeoff between helpfulness and safety may differ across use cases or change as policies evolve \citep{dong2023steerlm, wang2024conditioned}. In drug discovery, a medicinal chemist may want to explore different tradeoffs between binding affinity, solubility, and toxicity as a project progresses \citep{jin2020multi}. If models are aligned via reinforcement learning, every change to the reward function requires a new post-training run. 

Recent work \citet{frans2025diffusion} (CFGRL) shows that classifier-free guidance (CFG) \citep{ho2022classifier} can be used as a policy improvement operator by training a policy with a binary optimality conditioning and applying CFG to this binary conditioned policy at inference time. Their key result is that this guided distribution corresponds exactly to the solution of a policy improvement step. The conditional model implicitly encodes the advantage function, so no separate value network is needed.

We generalize this result 1) to autoregressive models and 2) beyond a binary optimality label to arbitrary reward functions that depend on a set of properties.

We consider a setting where each generated sequence $x$ has a vector of properties $y(x)$. We are given a reward function $r(y)$ that specifies how to weight these properties, and we want to steer generation toward high $r$ without retraining.

We study the following approach: at train time we construct a dataset of $(x,y)$ pairs and train a conditional model which produces next tokens as a function of current tokens and a fixed $y$ $\pi(x_{T} | y, x_{t<T})$. At inference time, given a new reward function $r$, perform \textbf{reward weighted classifier free guidance} (RCFG) by first sampling a \textbf{guidance set} $Y_{s}$ of $y$ then sampling using next token logits that are modified as follows: $$\pi_{RCFG} (x_{T} \mid x_{t<T}, r, \gamma) = \gamma \sum_{y \in Y_{S}} \hat{r}(y) (\pi(x_{T} \mid y, x_{t<T}) - (\pi(x_{T} \mid x_{t<T})) + \pi(x_{T} \mid x_{t<T})$$ where $\hat{r}(y)$ is $r(y)$ standardized within the sampled guidance set $Y_{S}$.

Our key theoretical insight is: \textbf{RCFG is an approximation to tilting the sampling distribution by the $Q$ function implied by the model and reward.} 

We apply RCFG to molecular generation and show that \textbf{RCFG can optimize novel reward functions at test time without retraining. RCFG can achieve a large fraction of the reward gains of full RL while maintaining generation diversity.}

Finally, we show that the RCFG policy can be thought of as privileged information (i.e. by combining the conditional policy and the known reward function). We then apply self-distillation \citep{penaloza2026privileged,zhao2026self} with the RCFG logits as the teacher and show that \textbf{RCFG warm starts can be used as a warm start to speed up RL for a given reward function.}

\section{Related Work}
The idea of steering generative models at inference time has a long history, from classifier guidance \citep{dhariwal2021diffusion, dathathri2020plug} and classifier-free guidance (CFG) \citep{ho2022classifier} in diffusion models to controlled decoding \citep{mudgal2024controlled, khanov2024args} and contrastive methods \citep{li2023contrastive, liu2021dexperts} in language models. CFG has also been applied directly to autoregressive LLMs by contrasting conditional and unconditional next-token distributions \citep{sanchez2023stay}. This paper shows that in the context of autoregressive models CFG can be thought of as an approximation to optimal tilting given the $Q$ function of the current policy and the reward.

\citet{rafailov2024from} show that a DPO-trained language model's logits define an optimal $Q$ function in the token-level MDP. Our derivation is related, though in our setting the given $r(y)$ directly provides the importance ratios to optimize rather than a labeled preference dataset.

The multi-objective conditioning approach is shared with SteerLM \citep{dong2023steerlm} and Rewards-in-Context \citep{yang2024rewards}. Both of these approaches train models which are distributions of the form $\pi(\text{output} \mid \text{input}, y)$. However, at inference time SteerLM simply uses the conditional model directly. This is an extremely useful approach when there are no tradeoffs between elements of $y$ or when the optimal tradeoff point $y^*$ is already known. However, in many cases a user may know their tradeoffs (for example, they are willing to trade $1$ unit of bioavailability for $.5$ units of lipophilicity or may want to maintain molecular weight at a given target while minimizing the number of hydrogen bond acceptors) but not the optimal set of properties $y^*$ that a molecule which optimizes these tradeoffs will have. This is precisely the use case for RCFG.

In our setting, like many of interest $r$ is defined over final sequences whereas steering needs to occur on partial sequences. Work like FUDGE \citep{yang2021fudge}, Controlled Decoding \citep{mudgal2024controlled}, or GenARM \citep{xu2024genarm} solve the per-token KL-regularized RL objective by learning a separate prefix scorer (value network) for partial sequences.  ARGS \citep{khanov2024args} gets a score for a partial sequence by applying a full sequence model directly. 

When using a method that outputs a scalar reward with a prefix scorer when reward functions change, a new prefix scorer needs to be trained. Our results show that a generative model conditioning on sampled values of $y$ can be combined with $r(y)$ to play the role of the prefix scorer at a cost of inference time compute.

Similarly, PAD \citep{chen2024pad} obtains token-level personalized rewards from the implicit Q function of a DPO-trained model without a separate value network. When preferences are linear combinations of learned latent features, PAD provides exact Q-function guidance. RCFG addresses the complementary setting where the user specifies an arbitrary reward function $r(y)$ over interpretable outcome attributes, requiring only a conditional generative model and no reward model training.

Rewards-in-Context (RiC) \citep{yang2024rewards} is a very closely related approach to ours. RiC also trains a single conditional model via supervised fine-tuning. At inference time, RiC maps a user's preference vector over reward dimensions to a single conditioning point $y^*$ and generates directly from the conditional model. 
 
In general, when $y^*$ is known ahead of time and well represented in the training data, it will be very hard to beat simple conditional generation on that set of properties. However, the optimal $y^*$ may lie in a sparsely supported region of property space, and \citet{yang2024rewards} find that even with online data augmentation, there are situations where the RiC-style approach can fail to produce a clear Pareto front (e.g. their Appendix C discusses the case of positively correlated $y$ where such failures can occur). 

\section{Formalization}
Let $\mathcal{X}$ be a set of finite length sequences with generic sequence $x$ and generic sequence element $x_T$. Let $\pi$ be a base distribution on $\mathcal{X}$ with autoregressive factorization:
$$\pi(x) = \prod_{T} \pi(x_{T} \mid x_{t<T}).$$
Assume that $\mathcal{X}$ is exhaustive, it is the set of all sequences that are reachable by our $\pi$.

Let $\mathcal{Y}$ be a set of descriptor vectors. For a full sequence $x$, let $y(x) \in \mathcal{Y}$ denote its properties. Assume that $y(x)$ is only defined on a subset of sequences but that with probability $1$ $\pi$ reaches such a sequence after a finite number of steps.

As a concrete example let $\mathcal{X}$ be the set of all molecules below a given size expressed in SMILES notation (e.g.\ CC(=O)Oc1ccccc1C(=O)O). Let $y$ be a vector of molecular properties (e.g.\ molecular weight, number of hydrogen bond acceptors, total polar surface area, etc.). $\pi(x \mid y)$ is a molecule generation model which inputs a set of property descriptors and autoregressively generates a SMILES string.

Given that $\pi$ is a distribution on $\mathcal{X}$ and $y$ is a deterministic function of $x$ we can slightly abuse notation and let $\pi(x \mid x_{t<T})$ be the probability of reaching the complete sequence $x$ given a current state $x_{t<T}$ and let $\pi(y \mid x_{t<T})$ be the distribution implied on $y$ given the current state.

Given a reward function $r(y)$, we can consider the following question. Suppose we have generated a sequence up to $x_{t<T}$, we are choosing what token to output $x_{T}$, afterwards we will continue with our policy $\pi.$ In other words, we are looking for the $Q$ function 
\begin{equation}\label{eq:q-def}
Q (x_T, x_{t<T}, r, \pi) = \sum_{y} r(y)\, \pi (y \mid x_{T}, x_{t<T}).
\end{equation}
 
Ideally, we would like to tilt the base policy by the Q-function values without going too far off the base policy. This means we want to sample from a tilted distribution $$\pi_{TILT} (x_T \mid x_{t<T}) \propto \pi(x_{T} \mid x_{t < T}) Q(x_T, x_{t<T}, r, \pi)^{\gamma}.$$

The logits of this distribution can be written as

$$\gamma (\log  Q (x_T, x_{t<T}, r, \pi)) + \log(\pi(x_T \mid x_{t<T})$$

Let us now look at the $Q$ function more directly. Applying Bayes' rule to $\pi(y \mid x_T, x_{t<T})$ in~\eqref{eq:q-def}:
\begin{equation}\label{eq:q-bayes}
Q_{\pi}(x_T, x_{t<T}) = \sum_{y} r(y)\, \pi(y \mid x_{t<T})\, \frac{\pi(x_T \mid y, x_{t<T})}{\pi(x_T \mid x_{t<T})}.
\end{equation}

The Q-function is an $r(y)\,\pi(y \mid x_{t<T})$-weighted sum of importance ratios, each of which is a classifier-free guidance term for outcome $y$~\citep{ho2022classifier}.

We now have several issues to make this fully tractable. First, autoregressive models work in log-space, where we have access to log-ratios. Let $\bar{r}(y)$ be the prior normalized reward. We can replace the ratio by the log ratio and then by Jensen's inequality:

\begin{equation}\label{eq:jensen}
\sum_{y} \bar{r}(y)\, \log \frac{\pi(x_T \mid y, x_{t<T}}{\pi(x_T \mid x_{t<T})} \;\leq\; \log \sum_{y} \bar{r}(y)\, \frac{\pi(x_T \mid y, x_{t<T})}{\pi(x_T \mid x_{t<T})}
\end{equation}

Thus the guidance vector $$\sum_{y} \bar{r}(y) (\log \pi(x_{T} \mid y, x_{t<T}) - \log \pi (x_{T} \mid x_{t<T}))$$ is a variational approximation to tilting by the true $(Q (x_{T}, x_{t<T}, r, \pi))$. 

Steering using RCFG moves mass toward high-$Q$ tokens but, due to the log, but underestimates the contribution of tokens with high variance across outcomes. 

This gives us the reward weighted CFG sampling with logits for tokens at time $t$ given by $$\gamma \sum_{y} \bar{r}(y) (\log \pi(x_{T} \mid y, x_{t<T}) - \log \pi (x_{T} \mid x_{t<T})) + \log \pi (x_{T} \mid x_{t<T})$$

\section{Practical Considerations}
There are several issues to computing RCFG in practice. First, computing the full sum over $y$ conditionals requires one forward pass per $y$, which can become expensive or impossible when $y$ is high-dimensional or continuous. We propose sampling a fixed set of $Y_{S}$ and only computing RCFG based on those. In the experiments we will vary the size of $Y_{S}$ while sampling $y$ from the distribution of the training data.

Second, the size of the step depends on the choice of $\gamma$ and arbitrary scaling of the reward function. We propose using the normalized reward rather than the raw reward $$A(y, Y_{S}) = \dfrac{r(y) - \mathrm{mean}_{Y_S}(r(y))}{\mathrm{std}_{Y_S}(r(y))}.$$ This gives a nice interpretation to $\gamma$ which states that a weight of $\gamma=1$ is trying to move the logits by one standard deviation of reward.

Third, the full approximation requires weighting by $\bar{r}(y) = \pi(y \mid x_{t<T}) r(y)$ which includes the distribution of $y$ that will be reached conditional on  $x_{t<T}$.

In practice, this could be computed via rollouts though this is computationally expensive. A much cheaper strategy is to sample $Y_{S}$ from our training data thus we will be weighting the importance ratios by the baseline distribution of $y$ instead of the $x_{t<T}$-conditional distribution of $y$. 

The sampled analog of our RCFG equation is $$\gamma \dfrac{1}{\mid Y_{S} \mid} \sum_{y \in Y_{S}} r(y) (\log \pi(x_{T} \mid y, x_{t<T}) - \log \pi (x_{T} \mid x_{t<T})) + \log \pi (x_{T} \mid x_{t<T})$$

Weighting by the prior has the effect of upweighting the RCFG contribution to rare tokens. In addition, our original Jensen bound is tightest when the importance ratios are well-behaved. The importance ratios are poorly behaved when the denominator in the ratio is small, allowing small absolute changes in the logit to become large proportional changes. 

For these two reasons, at each step we restrict our choice of $x_{T}$ after guidance to only tokens that appear in the $.95$ nucleus \citep{holtzman2019curious} of the unconditional $\pi ( \cdot, \mid x_{t<T})$.

Finally, recall the policy iteration algorithm. Given a reward function $r$ we begin with an arbitrary policy $\pi_0$. $Q$ learning updates the policy by updating the action at each state to be $$\pi_1(s) \leftarrow \text{argmax}_{a} Q(a, s, \pi_0, r).$$ Thus, tilting by the $Q$ function can be thought of, roughly, as trying to take one policy improvement step. 

We note that this is the same argument made by \cite{frans2025diffusion} in the binary $y$ case. In addition, we also see how the binary $y$ case is much simpler and indeed in this case CFG is a better approximation of $Q$. In the binary $y$ case RCFG collapses to weighting a single guidance vector $\gamma r(y=1) p(y = 1 \mid x_{t<T}).$ Here the conditional probability of $y$ is simply a rescaling of $r(y)$ since $r(y=0)$ can be set to be $0$. In the non-binary $y$ case, using the prior is not equivalent to rescaling since $r$ and the prior jointly weight many instances of $y$ simultaneously.

\section{Experiments}
Generating molecules that have certain properties is becoming an important applied topic of interest \citep{yang2024molgen,bagal2021molgpt}. Typically these works focus on a target properties $y^*$ and try to generate molecules that have these properties. We focus on a slightly different, related problem, where we are given a set of tradeoffs between properties (rather than a fixed $y^*$).

We take a large publicly available dataset of small drug like molecules. We use the canonical SMILES representation to transform each molecule into a unique string. Note that not every SMILES string is a valid molecule and so one of our metrics of interest is validity rate of produced strings. 

We use RDKit \citep{landrum2013rdkit} to extract $25$ properties that are optimization relevant from the molecules. For continuous properties (e.g. molecular weight) we bin them into discrete bins, for exact properties (e.g. number of hydrogen bond donators) or boolean ones (whether the molecule has a halogen) we use the exact count. See Table \ref{tab:properties} for a description.

For training a conditional $$\pi(\text{next token} \mid \text{properties, current tokens})$$ model we train a standard autoregressive LLM on a single string which is direct dump of the molecular properties dict concatenated to the molecule SMILES i.e.: $$\colorbox{gray!10}{\texttt{[BOS]Generate a molecule \{molecule\_config\}[SEP]\{SMILES string\}[EOS]}}.$$ 

We train with both complete and partial dropout of properties. With probability $.2$ we include the full config and with complementary probability we sample an integer $1$ to $n_{keys}$ and keep that many randomly selected keys in the config, dropping out the rest. 

We start with a pretrained language model, our initial explorations showed that these models already had some, but not perfect, understanding of the grammar of SMILES due to exposure in pre-training. We use Qwen3-0.6B-Base \citep{yang2025qwen3} as a starting point and continue training it using a very standard setup - HuggingFace transformers model class \citep{wolf2020transformers}, a learning rate of $3e-4$ annealed to $3e-5$ over the run with a cosine schedule and a $100$ step warmup, Adam optimizer \citep{kingma2014adam}, sequence packed batches of length $512$ and per-GPU batch size of $48$ yielding approximately $200,000$ tokens per step. We train for $50,000$ steps on a single $8 \times$ H100 node. The model sees approximately $10$ billion tokens. We do not epoch our dataset, every step sees unique molecules. 

This results in a strong base model: when conditioned on $1024$ examples of properties sampled from a set of molecules unseen at training the generated SMILES are valid molecules over $99$ percent of the time. Importantly, generated molecules agree with their input conditioning nearly perfectly - on average a one-shot generation satisfies more than $95 \%$ of input properties and a best-of-$16$ yields nearly $100 \%$ agreement. 

The base model maintains generation diversity, sampling generations from the model at temperature $.7$ yields a mean pairwise Tanimoto similarity of $\sim .15$, sampling $1000$ random pairs of molecules from the training data yields a Tanimoto similarity of $.14$.

We come up with $24$ reward functions that depend on between $1$ and $4$ of the config properties. We generated these by prompting Claude Opus to generate $50$ reward functions for this problem and then manually choose $24$ that appeared reasonable. These are chosen to be related to the kinds of tradeoffs medicinal chemists face in practice. Reward functions range from simple (maximize QED) to moderate (trade off between logP and polar surface area) to complex (keep molecular weight near $400$, TPSA near $90$, while minimizing number of hydrogen bond donators). See table \ref{tab:rewards} for all of our reward functions.

We choose the reward functions so that they are more complex than simple linear tradeoffs, so $y^*$ is not directly solveable via e.g. the linear program in \citep{yang2024rewards}, but our functions are simple enough that we can determine the $y^*$ which maximizes it by inspection. For functions which are unbounded in some variable, we clamp that variable to the $p90$ or $p10$ of our training data. 

In theory, one could exceed $r (y^*)$ by generating molecules whose properties fall far outside of our training dataset, which we find our RL able to do in two cases. However, there is also no guarantee that $r(y^*)$ is attainable given that it may require a combination of properties that is impossible in practice. Nevertheless, it is a an easy way to make scores comparable across reward functions and we normalize to $0$ being the expected reward of the baseline model, and $1$ being $r(y^*).$

We include several baselines. First, we include the oracle $\pi(\cdot \mid y^*)$. We also include the best-of-4 rejection sampling baseline. We also include an RL baseline. For each reward functions we first perform $1000$ steps of RL training on the base model with no config in the prompt. We use vanilla policy gradient with mean/std batch normalized advantages, a temperature of $.7$ and a batch size of $84$. We swept a partial grid of $16$ combinations of values of learning rates $[5e-6, 3e-5, 1e-4]$, KL regularization $[0, 0.01, 0.1]$ and values of entropy bonus $[0, 0.01, 0.025, 0.05]$ per reward function. We give a reward of $-1$ for generating an invalid SMILES string. We choose the best RL hyperparameters per reward function as the one that attains the highest reward without collapsing or diverging. We then run a longer, $2000$ step run with those hyperparameters. We use these as our baseline RL trained models.

For each reward function we also generate using RCFG. We sample $Y_{S}$ from real molecules held out from the training. Our main knob of interest is the dependence of outcomes on the size of $Y_{S}$ as this controls how many forward passes are required per token. 

We condition only on the keys of the config relevant to the reward function, since the model is trained with random config key dropout this remains on distribution for the base model. We swept $\gamma$ in $[1, 2, 4, 8]$ and found results extremely robust to values of $\gamma$ with some increase in reward between $\gamma=1$ and $\gamma=2$ but no changes after that, we report $\gamma=2$. As with the RL runs we use a sampling temperature of $0.7$.

Our hyperparameter sweeps mean we have a very strongly optimized RL baseline while for RCFG we use a set of hyperparameters (size of nucleus, how we sample $Y_{S}$, etc...) we found to work well in small amounts of testing. 

We see that there is substantial variance across reward functions, in particular for several of the unbounded functions the RL model achieves rewards above $r^*$ by generating legitimate molecules that are very much unlike anything in our training set. Whether this should be thought of as a form of reward hacking or legitimate optimization is unclear. To prevent this from coloring our results, we report performance both as the mean across reward functions and of the median reward function. 

We report aggregate results in the main text and in the Appendix include the results disaggregated by reward function. 

We see that RCFG achieves rewards comparable to many steps of RL without sacrificing diversity, suggesting that it is indeed a policy improvement operator. We see that, as expected, on average when $y^*$ is known and `valid' it is better to simply condition on that $y^*$. 

However, in the Appendix (table \ref{tab:method_comparison}) we do see that for $4$ of our reward functions, $\pi (\cdot \mid y^*)$ yields lower rewards than RL or RCFG. These reward functions are ones where we must trade off against two properties that are typically positively correlated and so the $y^*$ that is naively attained by setting one property high and another property low leads to a region of property space which is not well supported in the training data. This is precisely the failure mode of $y^*$ conditioning noted by RiC \citep{yang2024rewards}.

\begin{figure}
    \centering
    \includegraphics[width=1.0\linewidth]{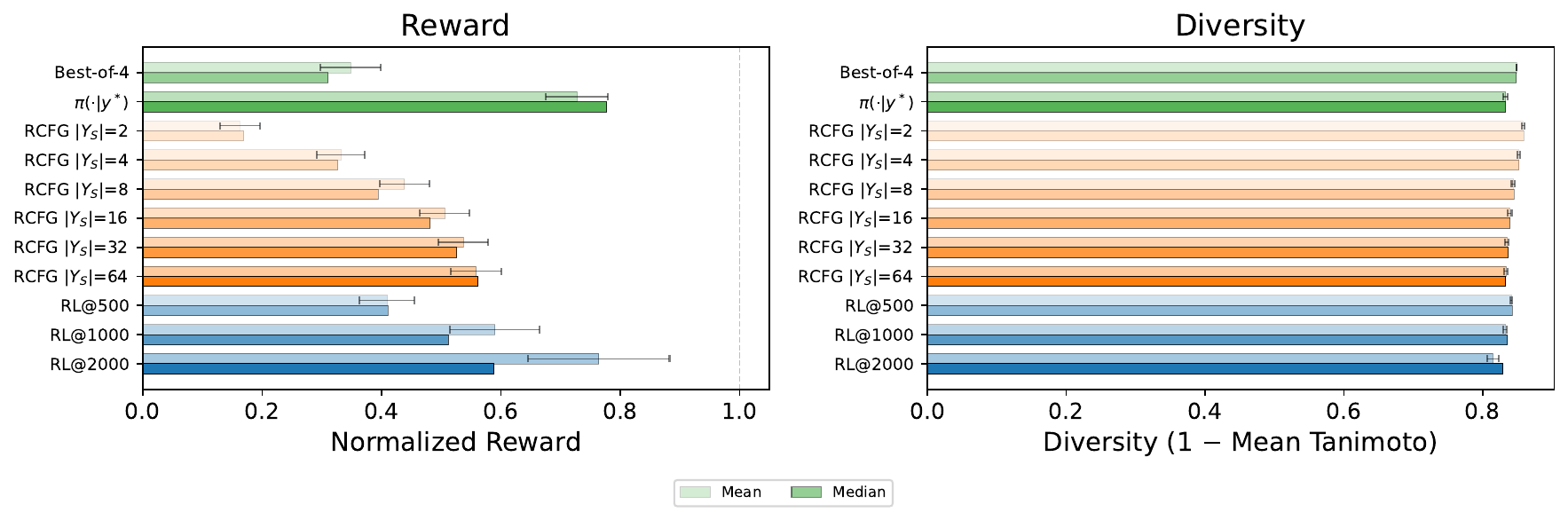}
    \caption{RCFG at inference time yields reward increases similar in magnitude to relatively large number of RL steps without sacrificing diversity. Results are shown both on average across all reward functions and for the median reward function. Error bars for the mean reflect standard errors.}
    \label{fig:rcfg_result}
\end{figure}

We also considered whether warm-starting RL with an RCFG-distilled policy could improve convergence speed. Before beginning RL training, we performed 50 steps of KL distillation using the RCFG-guided distribution as the teacher. Specifically, at each distillation step we generate a batch of molecules from the current policy $\pi_\theta$, then compute the RCFG teacher distribution $\pi_{\text{RCFG}}(\gamma=2, |Y_S|=8, r)$ using the frozen initial checkpoint. The loss minimizes the per-token forward KL divergence over response tokens:
$$\mathcal{L}_{\text{distill}} = \sum_{t} \text{KL}\left(\pi_{\text{RCFG}}(x_t \mid x_{<t}) || \pi_\theta(x_t \mid x_{<t})\right)$$
where the KL is computed over the full vocabulary distribution at each token position, not just the selected token. This distillation uses a separate optimizer (learning rate  $3e-5$) which is discarded after the warmup phase. RL then proceeds from the distilled checkpoint with a fresh optimizer using the same RL procedure as above.

This can be thought of a form of self-distillation where the reward function input and conditional policy plays the role of privileged information \citep{zhao2026self,penaloza2026privileged}.

\begin{figure}
    \centering
    \includegraphics[width=1\linewidth]{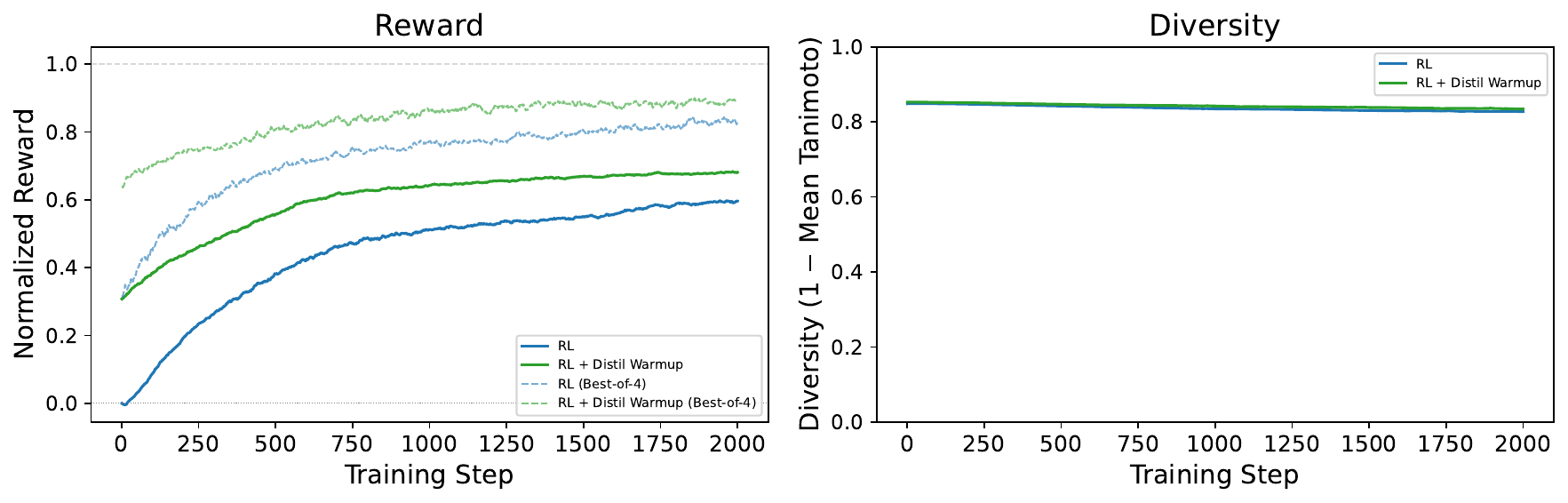}
    \caption{Using 50 steps of RCFG distillation as a warm start policy before beginning RL leads to much faster convergence as RCFG can be thought of as privileged information distillation. Plot shows median normalized at each step as several reward functions have extremely high normalized rewards skewing the average upward.}
    \label{fig:rl_result}
\end{figure}

We found that this works well to improve RL convergence speed in our setting while still not giving up generation diversity. Figure \ref{fig:rl_result} shows achieved reward of the median reward function by RL step. In addition, we see that we are not simply ``trading pass@N for pass@1'' \citep{research2026composer}. Rather, as we RL for longer the best-of-4 accuracy for that policy goes up as well.

We tried to build on this result. We tried longer periods of RCFG distillation, but found that the policy did not improve much beyond the first set of steps and then eventually diverged completely. 

We also tried performing RCFG distillation multiple times during the RL run - start with a base policy $\pi^0$, distil towards $\pi^0_{RCFG}$, perform $500$ steps of RL to get $\pi^1$ from $\pi^0_{RCFG}$, try to use RCFG again on top of $\pi^1$, etc... 

We found that this strategy worked on a subset of reward functions for early steps ($<1000$) but was quite unstable in general. We believe that this comes from the fact that the conditional model $\pi^1(x \mid y)$ becomes stale relative to the kinds of molecules that $\pi^1$ generates naturally. Thus, combining RCFG with RL beyond simply using it as a warm start likely requires some form of continued SFT on the conditional policy or other tricks and we leave this question for future research.

\section{Conclusion, Limitations, and Future Work}
We have presented RCFG, a simple method for inference time policy improvement in autoregressive language models. We have show than RCFG is strongly related to tilting by the $Q$ function. We have shown that RCFG works in the case of molecular property optimization and can be helpful both as an inference time policy improvement and as a warm start for RL.

We focused on molecular generation because it is both genuinely interesting and important and because we found that standard LLM preference datasets mostly leave little room for meaningful tradeoff optimization as most attributes are somewhat independent (e.g. in SteerLM attributes like helpful and funny are mostly independent so there is no tradeoff) or are rare (e.g. most responses are not harmful). Molecular properties, by contrast, exhibit genuine conflicts that require navigating a non-trivial Pareto frontier. Expanding RCFG to other tasks is an extremely interesting direction for future work.

\section{LLM Usage Statement}
Claude Opus 4.6 was used heavily by the authors in the production of this research. The model was used for code generation, proofreading of the manuscript, to help authors find related work, and for acting like a `extremely critical but scientifically fair referee' to suggest ways of strengthening the experimental baselines as well as our exposition and framing with respect to existing literature. All errors whether of omission or commission are the responsibility of the human authors.

\clearpage 

\begin{table}[t]
\centering
\caption{Molecular properties used as conditioning inputs. Continuous properties are discretized into fixed-width bins (\texttt{min}/\texttt{max} interval); integer properties use exact values; binary properties use boolean flags; and list properties enumerate present items.}
\label{tab:properties}
\tiny
\begin{tabular}{lll}
\toprule
\textbf{Config Key} & \textbf{Description} & \textbf{Discretization} \\
\midrule
\multicolumn{3}{l}{\textit{Physicochemical (continuous $\rightarrow$ interval)}} \\
\texttt{mw}           & Molecular weight (Da)              & Bins of width 25 \\
\texttt{logp}         & Crippen LogP                       & Bins of width 0.5 \\
\texttt{tpsa}         & Topological polar surface area     & Bins of width 10 \\
\texttt{mr}           & Molar refractivity                 & Bins of width 10 \\
\texttt{fsp3}         & Fraction of sp$^3$ carbons         & Bins of width 0.1 \\
\texttt{qed}          & Quantitative estimate of druglikeness & Bins of width 0.1 \\
\texttt{sa\_score}    & Synthetic accessibility score      & Bins of width 0.5 \\
\midrule
\multicolumn{3}{l}{\textit{Counts (exact integer)}} \\
\texttt{hbd}          & Hydrogen-bond donors               & Exact \\
\texttt{hba}          & Hydrogen-bond acceptors            & Exact \\
\texttt{rotb}         & Rotatable bonds                    & Exact \\
\texttt{rings}        & Total ring count                   & Exact \\
\texttt{arom\_rings}  & Aromatic ring count                & Exact \\
\texttt{stereo}       & Stereocenters                      & Exact \\
\texttt{heteroatoms}  & Heteroatom count                   & Exact \\
\texttt{heavy\_atoms} & Heavy atom count                   & Exact \\
\texttt{max\_ring\_size} & Largest ring size               & Exact \\
\texttt{charge}       & Formal charge                      & Exact \\
\texttt{lipinski\_violations} & Lipinski rule-of-5 violations & Exact \\
\midrule
\multicolumn{3}{l}{\textit{Binary flags}} \\
\texttt{has\_basic\_n}   & Contains basic nitrogen          & Boolean \\
\texttt{has\_acid}       & Contains acidic group            & Boolean \\
\texttt{has\_halogen}    & Contains any halogen             & Boolean \\
\texttt{has\_spiro}      & Contains spiro center            & Boolean \\
\texttt{has\_bridgehead} & Contains bridgehead atom         & Boolean \\
\midrule
\multicolumn{3}{l}{\textit{Lists (present items enumerated)}} \\
\texttt{halogens}     & Which halogens are present (F, Cl, Br, I) & List of present elements \\
\texttt{scaffolds}    & Ring systems (e.g.\ piperidine, indole) & List of matched SMARTS \\
\texttt{func\_groups} & Functional groups (e.g.\ amide, ester) & List of nonzero fragment counts \\
\bottomrule
\end{tabular}
\caption{Extracted properties for conditional model training.}

\end{table}

\begin{table}[t]
\centering
\caption{Reward functions used for guided generation. Each operates on the discretized config and returns a scalar score. $m(\cdot)$ denotes the bin midpoint for interval-valued keys.}
\label{tab:rewards}
\tiny
\begin{tabular}{llp{5.8cm}}
\toprule
\textbf{Name} & \textbf{Formula} & \textbf{Description} \\
\midrule
\texttt{maximize\_qed} & $m(\text{qed})$ & Maximizes drug-likeness score \\
\texttt{drug\_like} & $2\,\text{qed} + \triangle(\text{logP},2.5) + \triangle(\text{MW},350) - 0.5\,v$ & Lipinski-friendly, high QED, moderate logP \\
\texttt{high\_mw\_high\_qed} & $\text{qed} + \max(0, \text{MW}-300)/200$ & High molecular weight AND high QED \\
\texttt{low\_tpsa\_high\_logp} & $\text{logP} - \text{TPSA}/60$ & High membrane permeability \\
\texttt{cns\_penetrant} & $\triangle(\text{MW},350) + (1-\text{TPSA}/90)^+ + \triangle(\text{logP},2) - 0.3\,\text{HBD}$ & CNS-penetrating drug profile \\
\texttt{oral\_bioavailable} & $\triangle(\text{MW},400) + \triangle(\text{logP},2.5) + (1-\text{rotb}/10)^+ - v$ & Oral bioavailability (Lipinski + low flexibility) \\
\texttt{fragment\_like} & $\triangle(\text{MW},200) + \triangle(\text{logP},1) + \triangle(\text{rings},2)$ & Fragment-sized molecule (Rule of Three) \\
\texttt{lead\_like} & $\triangle(\text{MW},300) + \triangle(\text{logP},1) + (1-\text{rotb}/7)^+ + (1-\text{HBD}/3)^+$ & Lead-like chemical space \\
\texttt{ppi\_inhibitor} & $(\text{MW}-350)^+\!/200 + 2\,\text{Fsp3} + (\text{rings}-2)^+\!/3 + \triangle(\text{logP},3)$ & Protein--protein interaction inhibitor \\
\texttt{kinase\_like} & $\triangle(\text{arom},2.5) + \triangle(\text{HBA},5) + \triangle(\text{MW},450) + \triangle(\text{HBD},2)$ & Kinase inhibitor profile \\
\texttt{soluble\_permeable} & $\triangle(\text{logP},2) + \triangle(\text{TPSA},70) + \triangle(\text{MW},350)$ & Balanced solubility/permeability \\
\texttt{metab\_stable} & $3\,\text{Fsp3} - 0.4\,\text{arom\_rings} + \triangle(\text{MW},380)$ & Metabolic stability (high sp$^3$) \\
\texttt{easy\_synth\_druglike} & $2\,\text{qed} + (4-\text{SA})^+\!/3$ & Synthesizable AND drug-like \\
\texttt{macrocycle\_like} & $(\text{max\_ring}-8)^+\!/5 + (\text{MW}-400)^+\!/200 + (1-\text{rotb}/8)^+ + \triangle(\text{TPSA},100)$ & Macrocyclic compound \\
\texttt{gpcr\_like} & $\triangle(\text{MW},400) + \triangle(\text{logP},2.5) + \triangle(\text{rings},3) + \mathbb{1}[\text{basic\_N}]$ & GPCR ligand profile \\
\texttt{low\_tox\_proxy} & $(1-\text{logP}/4)^+ + \triangle(\text{MW},350) - \mathbb{1}[\text{halogen}]$ & Low-toxicity proxy (no halogens, low logP) \\
\texttt{3d\_complex} & $2\,\text{Fsp3} + 0.5\min(\text{stereo},4) + (\text{rings}-1)^+\!/4 + 0.5\,\mathbb{1}[\text{bridge}]$ & Three-dimensional complexity \\
\texttt{antibacterial} & $(\text{TPSA}-60)^+\!/80 + (1-\text{MW}/500)^+ + 0.3\min(\text{HBD},4) + \triangle(\text{logP},0)$ & Gram-negative antibacterial profile \\
\texttt{lipophilic\_eff} & $2\,\text{qed} - 0.5(\text{logP}-1)^+$ & High QED without lipophilicity \\
\texttt{rigid\_potent} & $(1-\text{rotb}/8)^+ + (\text{rings}-1)^+\!/3 + \triangle(\text{MW},380)$ & Rigid scaffold with multiple rings \\
\texttt{topical\_drug} & $\triangle(\text{logP},3.5) + (1-\text{MW}/500)^+ + \triangle(\text{TPSA},50)$ & Topical/dermal drug profile \\
\texttt{inhaled\_drug} & $(1-\text{MW}/500)^+ + \triangle(\text{logP},2) + (1-\text{TPSA}/100)^+ + (1-\text{rotb}/8)^+$ & Inhaled drug profile \\
\texttt{sel\_3d\_druglike} & $2\,\text{Fsp3} + 0.4\min(\text{stereo},3) + 1.5\,\text{qed}$ & Selective \& drug-like (3D + high QED) \\
\texttt{low\_mw\_potent} & $(1-\text{MW}/400)^+ + 1.5\,\text{qed} + (\text{HBA}+\text{HBD}-2)^+\!/6$ & Small but potent (low MW + high QED) \\
\bottomrule
\end{tabular}
\vspace{0.3em}
\par\noindent\footnotesize{$\triangle(x,t)$~$= \max\!\big(0,\; 1 - |x-t|/w\big)$ is a triangular kernel centered at target $t$ with task-specific width $w$; $({\cdot})^+ = \max(0,\cdot)$; $v$ = Lipinski violations; $\mathbb{1}[\cdot]$ = indicator function.}
\end{table}

\clearpage

\bibliography{main.bbl}

\begin{thebibliography}{32}
\providecommand{\natexlab}[1]{#1}
\providecommand{\url}[1]{\texttt{#1}}
\expandafter\ifx\csname urlstyle\endcsname\relax
  \providecommand{\doi}[1]{doi: #1}\else
  \providecommand{\doi}{doi: \begingroup \urlstyle{rm}\Url}\fi

\bibitem[Bagal et~al.(2021)Bagal, Aggarwal, Vinod, and
  Priyakumar]{bagal2021molgpt}
Viraj Bagal, Rishal Aggarwal, PK~Vinod, and U~Deva Priyakumar.
\newblock Molgpt: molecular generation using a transformer-decoder model.
\newblock \emph{Journal of chemical information and modeling}, 62\penalty0
  (9):\penalty0 2064--2076, 2021.

\bibitem[Chen et~al.(2024)Chen, Zhang, Luo, Chai, and Liu]{chen2024pad}
Ruizhe Chen, Xiaotian Zhang, Meng Luo, Wenhao Chai, and Zuozhu Liu.
\newblock Pad: Personalized alignment of llms at decoding-time.
\newblock \emph{arXiv preprint arXiv:2410.04070}, 2024.

\bibitem[Dathathri et~al.(2020)Dathathri, Madotto, Lan, Hung, Frank, Molino,
  Yosinski, and Liu]{dathathri2020plug}
Sumanth Dathathri, Andrea Madotto, Janice Lan, Jane Hung, Eric Frank, Piero
  Molino, Jason Yosinski, and Rosanne Liu.
\newblock Plug and play language models: A simple approach to controlled text
  generation.
\newblock In \emph{International Conference on Learning Representations}, 2020.

\bibitem[Dhariwal \& Nichol(2021)Dhariwal and Nichol]{dhariwal2021diffusion}
Prafulla Dhariwal and Alexander Nichol.
\newblock Diffusion models beat {GANs} on image synthesis.
\newblock In \emph{Advances in Neural Information Processing Systems},
  volume~34, pp.\  8780--8794, 2021.

\bibitem[Dong et~al.(2023)Dong, Wang, Sreedhar, Wu, and
  Kuchaiev]{dong2023steerlm}
Yi~Dong, Zhilin Wang, Makesh~Narsimhan Sreedhar, Xianchao Wu, and Oleksii
  Kuchaiev.
\newblock {SteerLM}: Attribute conditioned {SFT} as an (user-steerable)
  alternative to {RLHF}.
\newblock In \emph{Findings of the Association for Computational Linguistics:
  EMNLP 2023}, pp.\  11275--11288, 2023.

\bibitem[Frans et~al.(2025)Frans, Park, Abbeel, and Levine]{frans2025diffusion}
Kevin Frans, Seohong Park, Pieter Abbeel, and Sergey Levine.
\newblock Diffusion guidance is a controllable policy improvement operator.
\newblock \emph{arXiv preprint arXiv:2505.23458}, 2025.

\bibitem[Ho \& Salimans(2022)Ho and Salimans]{ho2022classifier}
Jonathan Ho and Tim Salimans.
\newblock Classifier-free diffusion guidance.
\newblock \emph{arXiv preprint arXiv:2207.12598}, 2022.

\bibitem[Holtzman et~al.(2019)Holtzman, Buys, Du, Forbes, and
  Choi]{holtzman2019curious}
Ari Holtzman, Jan Buys, Li~Du, Maxwell Forbes, and Yejin Choi.
\newblock The curious case of neural text degeneration.
\newblock \emph{arXiv preprint arXiv:1904.09751}, 2019.

\bibitem[Jin et~al.(2020)Jin, Barzilay, and Jaakkola]{jin2020multi}
Wengong Jin, Regina Barzilay, and Tommi Jaakkola.
\newblock Multi-objective molecule generation using interpretable
  substructures.
\newblock In \emph{Proceedings of the 37th International Conference on Machine
  Learning}, pp.\  4849--4859, 2020.

\bibitem[Khanov et~al.(2024)Khanov, Burapacheep, and Li]{khanov2024args}
Maxim Khanov, Jirayu Burapacheep, and Yixuan Li.
\newblock {ARGS}: Alignment as reward-guided search.
\newblock In \emph{International Conference on Learning Representations}, 2024.

\bibitem[Kingma \& Ba(2014)Kingma and Ba]{kingma2014adam}
Diederik~P Kingma and Jimmy Ba.
\newblock Adam: A method for stochastic optimization.
\newblock \emph{arXiv preprint arXiv:1412.6980}, 2014.

\bibitem[Landrum et~al.(2013)]{landrum2013rdkit}
Greg Landrum et~al.
\newblock Rdkit documentation.
\newblock \emph{Release}, 1\penalty0 (1-79):\penalty0 4, 2013.

\bibitem[Li et~al.(2023)Li, Holtzman, Fried, Liang, Eisner, Hashimoto,
  Zettlemoyer, and Lewis]{li2023contrastive}
Xiang~Lisa Li, Ari Holtzman, Daniel Fried, Percy Liang, Jason Eisner, Tatsunori
  Hashimoto, Luke Zettlemoyer, and Mike Lewis.
\newblock Contrastive decoding: Open-ended text generation as optimization.
\newblock In \emph{Proceedings of the 61st Annual Meeting of the Association
  for Computational Linguistics}, 2023.

\bibitem[Liu et~al.(2021)Liu, Sap, Lu, Swayamdipta, Bhagavatula, Smith, and
  Choi]{liu2021dexperts}
Alisa Liu, Maarten Sap, Ximing Lu, Swabha Swayamdipta, Chandra Bhagavatula,
  Noah~A Smith, and Yejin Choi.
\newblock {DExperts}: Decoding-time controlled text generation with experts and
  anti-experts.
\newblock In \emph{Proceedings of the 59th Annual Meeting of the Association
  for Computational Linguistics}, 2021.

\bibitem[Mudgal et~al.(2024)Mudgal, Lee, Ganapathy, Li, Wang, Huang, Chen,
  Cheng, Collins, Strohman, Chen, Beutel, and Beirami]{mudgal2024controlled}
Sidharth Mudgal, Jong Lee, Harish Ganapathy, YaGuang Li, Tao Wang, Yanping
  Huang, Zhifeng Chen, Heng-Tze Cheng, Michael Collins, Trevor Strohman, Jilin
  Chen, Alex Beutel, and Ahmad Beirami.
\newblock Controlled decoding from language models.
\newblock In \emph{Proceedings of the 41st International Conference on Machine
  Learning}, 2024.

\bibitem[Ouyang et~al.(2022)Ouyang, Wu, Jiang, Almeida, Wainwright, Mishkin,
  Zhang, Agarwal, Slama, Ray, et~al.]{ouyang2022training}
Long Ouyang, Jeffrey Wu, Xu~Jiang, Diogo Almeida, Carroll Wainwright, Pamela
  Mishkin, Chong Zhang, Sandhini Agarwal, Katarina Slama, Alex Ray, et~al.
\newblock Training language models to follow instructions with human feedback.
\newblock \emph{Advances in Neural Information Processing Systems},
  35:\penalty0 27730--27744, 2022.

\bibitem[Penaloza et~al.(2026)Penaloza, Vattikonda, Gontier, Lacoste, Charlin,
  and Caccia]{penaloza2026privileged}
Emiliano Penaloza, Dheeraj Vattikonda, Nicolas Gontier, Alexandre Lacoste,
  Laurent Charlin, and Massimo Caccia.
\newblock Privileged information distillation for language models.
\newblock \emph{arXiv preprint arXiv:2602.04942}, 2026.

\bibitem[Rafailov et~al.(2023)Rafailov, Sharma, Mitchell, Ermon, Manning, and
  Finn]{rafailov2023direct}
Rafael Rafailov, Archit Sharma, Eric Mitchell, Stefano Ermon, Christopher~D
  Manning, and Chelsea Finn.
\newblock Direct preference optimization: Your language model is secretly a
  reward model.
\newblock \emph{Advances in Neural Information Processing Systems}, 36, 2023.

\bibitem[Rafailov et~al.(2024)Rafailov, Hejna, Park, and
  Finn]{rafailov2024from}
Rafael Rafailov, Joey Hejna, Ryan Park, and Chelsea Finn.
\newblock From $r$ to $q^*$: Your language model is secretly a q-function.
\newblock \emph{arXiv preprint arXiv:2404.12358}, 2024.
\newblock COLM 2024.

\bibitem[Research et~al.(2026)Research, Chan, Shalaby, Wettig, Sanger, Zhai,
  Ajay, Nair, Snell, Lu, et~al.]{research2026composer}
Cursor Research, Aaron Chan, Ahmed Shalaby, Alexander Wettig, Aman Sanger,
  Andrew Zhai, Anurag Ajay, Ashvin Nair, Charlie Snell, Chen Lu, et~al.
\newblock Composer 2 technical report.
\newblock \emph{arXiv preprint arXiv:2603.24477}, 2026.

\bibitem[Sanchez et~al.(2024)Sanchez, Fan, Spangher, Levi, Ammanamanchi, and
  Biderman]{sanchez2023stay}
Guillaume Sanchez, Honglu Fan, Alexander Spangher, Elad Levi, Pawan~Sasanka
  Ammanamanchi, and Stella Biderman.
\newblock Stay on topic with classifier-free guidance.
\newblock In \emph{Proceedings of the 41st International Conference on Machine
  Learning}, 2024.

\bibitem[Schulman et~al.(2017)Schulman, Wolski, Dhariwal, Radford, and
  Klimov]{schulman2017proximal}
John Schulman, Filip Wolski, Prafulla Dhariwal, Alec Radford, and Oleg Klimov.
\newblock Proximal policy optimization algorithms.
\newblock \emph{arXiv preprint arXiv:1707.06347}, 2017.

\bibitem[Shao et~al.(2024)Shao, Wang, Zhu, Xu, Song, Zhang, Li, Wu, and
  Guo]{shao2024deepseekmath}
Zhihong Shao, Peiyi Wang, Qihao Zhu, Runxin Xu, Junxiao Song, Mingchuan Zhang,
  Y.~K. Li, Y.~Wu, and Daya Guo.
\newblock {DeepSeekMath}: Pushing the limits of mathematical reasoning in open
  language models.
\newblock \emph{arXiv preprint arXiv:2402.03300}, 2024.

\bibitem[Wang et~al.(2024)Wang, Kidambi, Sullivan, Agarwal, Dann, Szepesvari,
  and Joachims]{wang2024conditioned}
Kaiwen Wang, Rahul Kidambi, Ryan Sullivan, Alekh Agarwal, Christoph Dann,
  Andrea Szepesvari, and Thorsten Joachims.
\newblock Conditioned language policy: A general framework for steerable
  multi-objective finetuning.
\newblock In \emph{Findings of the Association for Computational Linguistics:
  EMNLP 2024}, 2024.

\bibitem[Wolf et~al.(2020)Wolf, Debut, Sanh, Chaumond, Delangue, Moi, Cistac,
  Rault, Louf, Funtowicz, et~al.]{wolf2020transformers}
Thomas Wolf, Lysandre Debut, Victor Sanh, Julien Chaumond, Clement Delangue,
  Anthony Moi, Pierric Cistac, Tim Rault, R{\'e}mi Louf, Morgan Funtowicz,
  et~al.
\newblock Transformers: State-of-the-art natural language processing.
\newblock In \emph{Proceedings of the 2020 conference on empirical methods in
  natural language processing: system demonstrations}, pp.\  38--45, 2020.

\bibitem[Xu et~al.(2024)Xu, Sehwag, Koppel, Zhu, An, Huang, and
  Ganesh]{xu2024genarm}
Yuancheng Xu, Udari~Madhushani Sehwag, Alec Koppel, Sicheng Zhu, Bang An,
  Furong Huang, and Sumitra Ganesh.
\newblock Genarm: Reward guided generation with autoregressive reward model for
  test-time alignment.
\newblock \emph{arXiv preprint arXiv:2410.08193}, 2024.

\bibitem[Yang et~al.(2025)Yang, Li, Yang, Zhang, Hui, Zheng, Yu, Gao, Huang,
  Lv, et~al.]{yang2025qwen3}
An~Yang, Anfeng Li, Baosong Yang, Beichen Zhang, Binyuan Hui, Bo~Zheng, Bowen
  Yu, Chang Gao, Chengen Huang, Chenxu Lv, et~al.
\newblock Qwen3 technical report.
\newblock \emph{arXiv preprint arXiv:2505.09388}, 2025.

\bibitem[Yang et~al.(2024{\natexlab{a}})Yang, Duke, Sornberger, Ogbaje, Risko,
  and Ganapathysubramanian]{yang2024molgen}
Chih-Hsuan Yang, Rebekah Duke, Parker~Delaney Sornberger, Moses Ogbaje, Chad
  Risko, and Baskar Ganapathysubramanian.
\newblock Molgen-transformer: An open-source self-supervised model for
  molecular generation and latent space exploration.
\newblock In \emph{AI for Accelerated Materials Design-NeurIPS 2024},
  2024{\natexlab{a}}.

\bibitem[Yang \& Klein(2021)Yang and Klein]{yang2021fudge}
Kevin Yang and Dan Klein.
\newblock Fudge: Controlled text generation with future discriminators.
\newblock In \emph{Proceedings of the 2021 Conference of the North American
  Chapter of the Association for Computational Linguistics: Human Language
  Technologies}, pp.\  3511--3535, 2021.

\bibitem[Yang et~al.(2024{\natexlab{b}})Yang, Pan, Luo, Qiu, Zhong, Yu, and
  Chen]{yang2024rewards}
Rui Yang, Xiaoman Pan, Feng Luo, Shuang Qiu, Han Zhong, Dong Yu, and Jianshu
  Chen.
\newblock Rewards-in-context: Multi-objective alignment of foundation models
  with dynamic preference adjustment.
\newblock In \emph{Proceedings of the 41st International Conference on Machine
  Learning}, 2024{\natexlab{b}}.

\bibitem[Zhao et~al.(2026)Zhao, Xie, Liu, Huang, Pang, Chen, and
  Grover]{zhao2026self}
Siyan Zhao, Zhihui Xie, Mengchen Liu, Jing Huang, Guan Pang, Feiyu Chen, and
  Aditya Grover.
\newblock Self-distilled reasoner: On-policy self-distillation for large
  language models.
\newblock \emph{arXiv preprint arXiv:2601.18734}, 2026.

\bibitem[Ziegler et~al.(2019)Ziegler, Stiennon, Wu, Brown, Radford, Amodei,
  Christiano, and Irving]{ziegler2019fine}
Daniel~M Ziegler, Nisan Stiennon, Jeffrey Wu, Tom~B Brown, Alec Radford, Dario
  Amodei, Paul Christiano, and Geoffrey Irving.
\newblock Fine-tuning language models from human preferences.
\newblock \emph{arXiv preprint arXiv:1909.08593}, 2019.

\end{thebibliography}

\bibliographystyle{colm2026_conference}

\clearpage 

\appendix
\section{Appendix}

\begin{table}[h]
\centering
\tiny
\begin{tabular}{lccccccc||ccc}
\toprule
 & \multicolumn{7}{c||}{Inference-time} & \multicolumn{3}{c}{RL} \\
Reward Function & $\pi(\cdot | y^*)$ & $|Y_S|$=2 & $|Y_S|$=4 & $|Y_S|$=8 & $|Y_S|$=16 & $|Y_S|$=32 & $|Y_S|$=64 & RL@500 & RL@1000 & RL@2000 \\
\midrule
3d\_complex & \textbf{0.97} & 0.19 & 0.39 & 0.53 & 0.62 & 0.63 & 0.68 & 0.73 & 0.90 & 0.92 \\
antibacterial\_like & \textbf{0.90} & 0.30 & 0.48 & 0.60 & 0.67 & 0.66 & 0.69 & 0.58 & 0.79 & 0.84 \\
cns\_penetrant & \textbf{0.52} & 0.04 & 0.23 & 0.33 & 0.43 & 0.47 & 0.48 & 0.42 & 0.50 & 0.57 \\
drug\_like & \textbf{0.71} & -0.14 & 0.04 & 0.21 & 0.32 & 0.39 & 0.45 & 0.25 & 0.26 & 0.39 \\
easy\_synth\_druglike & \textbf{0.86} & 0.24 & 0.40 & 0.48 & 0.56 & 0.58 & 0.62 & 0.40 & 0.60 & 0.99 \\
fragment\_like & \textbf{0.71} & 0.05 & 0.16 & 0.25 & 0.29 & 0.32 & 0.34 & 0.10 & 0.11 & 0.11 \\
gpcr\_like & \textbf{0.85} & 0.02 & 0.22 & 0.34 & 0.42 & 0.53 & 0.54 & 0.10 & 0.32 & 0.49 \\
high\_mw\_high\_qed & \textbf{0.70} & 0.34 & 0.47 & 0.55 & 0.58 & 0.58 & 0.59 & 0.41 & 1.26 & 2.84 \\
inhaled\_drug & \textbf{0.56} & -0.04 & 0.06 & 0.17 & 0.26 & 0.28 & 0.29 & 0.24 & 0.38 & 0.51 \\
kinase\_like & \textbf{0.92} & 0.13 & 0.29 & 0.40 & 0.51 & 0.52 & 0.57 & 0.32 & 0.47 & 0.48 \\
lead\_like & \textbf{0.84} & 0.08 & 0.19 & 0.27 & 0.35 & 0.38 & 0.41 & 0.27 & 0.45 & 0.51 \\
lipophilic\_efficiency & -0.14 & 0.27 & 0.47 & 0.56 & 0.60 & \textbf{0.60} & 0.59 & 0.44 & 0.55 & 0.59 \\
low\_mw\_high\_potency & 0.41 & 0.15 & 0.30 & 0.39 & \textbf{0.45} & 0.44 & 0.44 & 0.30 & 0.41 & 0.47 \\
low\_tox\_proxy & 0.54 & 0.20 & 0.55 & 0.74 & 0.77 & 0.82 & \textbf{0.86} & 0.63 & 0.96 & 1.14 \\
low\_tpsa\_high\_logp & 0.89 & 0.51 & 0.79 & 0.92 & 1.05 & 1.13 & \textbf{1.16} & 0.91 & 1.76 & 2.10 \\
macrocycle\_like & \textbf{1.23} & 0.08 & 0.13 & 0.17 & 0.18 & 0.21 & 0.21 & 0.14 & 0.25 & 0.77 \\
maximize\_qed & \textbf{0.74} & 0.40 & 0.54 & 0.62 & 0.60 & 0.59 & 0.60 & 0.43 & 0.52 & 0.58 \\
metabolically\_stable & \textbf{0.96} & 0.35 & 0.61 & 0.77 & 0.83 & 0.89 & 0.92 & 0.78 & 0.88 & 0.95 \\
oral\_bioavailable & \textbf{0.56} & -0.03 & 0.11 & 0.20 & 0.32 & 0.36 & 0.35 & 0.16 & 0.26 & 0.33 \\
ppi\_inhibitor & \textbf{0.86} & 0.13 & 0.23 & 0.36 & 0.46 & 0.49 & 0.55 & 0.49 & 0.62 & 0.70 \\
rigid\_potent & \textbf{0.82} & 0.34 & 0.48 & 0.53 & 0.56 & 0.58 & 0.59 & 0.55 & 0.58 & 0.62 \\
selective\_3d\_druglike & \textbf{0.81} & 0.24 & 0.46 & 0.59 & 0.67 & 0.67 & 0.68 & 0.69 & 0.79 & 0.80 \\
soluble\_permeable & \textbf{0.60} & -0.15 & -0.00 & 0.14 & 0.25 & 0.35 & 0.39 & 0.03 & 0.05 & 0.04 \\
topical\_drug & \textbf{0.63} & 0.21 & 0.35 & 0.39 & 0.40 & 0.41 & 0.41 & 0.47 & 0.51 & 0.57 \\
\midrule
Mean & 0.73 & 0.16 & 0.33 & 0.44 & 0.51 & 0.54 & 0.56 & 0.41 & 0.59 & 0.76 \\
Median & 0.78 & 0.17 & 0.33 & 0.39 & 0.48 & 0.53 & 0.56 & 0.41 & 0.51 & 0.59 \\
\bottomrule
\end{tabular}
\caption{Normalized reward across methods and reward functions. Scores normalized so 0 = baseline model, 1 = $r(y^*)$. Best per row in \textbf{bold}.}
\label{tab:method_comparison}
\end{table}

\end{document}